\documentclass[journal,web,twoside]{ieeecolor} 
\usepackage{generic}
\usepackage{amsmath,amssymb,amsfonts}
\usepackage{algorithmic}
\usepackage{cite}
\usepackage{graphicx}
\usepackage{algorithm,algorithmic}
\usepackage{hyperref}
\usepackage{soul}
\hypersetup{colorlinks=true,linkcolor=blue,citecolor=blue,urlcolor=blue}
\usepackage{textcomp}
\usepackage{caption}
\usepackage{multirow}
\def\BibTeX{{\rm B\kern-.05em{\sc i\kern-.025em b}\kern-.08em
    T\kern-.1667em\lower.7ex\hbox{E}\kern-.125emX}}
\begin{document}



\begin{center}
    \Large  
    \textbf{IEEE COPYRIGHT NOTICE}
    
    \vspace{2cm}  
    
    \normalsize  
    \parbox{0.8\textwidth}{  
        \textbf{© 2024 IEEE.  Personal use of this material is permitted.  Permission from IEEE must be obtained for all other uses, in any current or future media, including reprinting/republishing this material for advertising or promotional purposes, creating new collective works, for resale or redistribution to servers or lists, or reuse of any copyrighted component of this work in other works.}
    }
\end{center}

\markboth{\hskip25pc IEEE Journal of Biomedical and Health Informatics}
{CS Tung \MakeLowercase{\textit{et al.}}: A Hybrid AI System for Automated EEG Background Analysis and Report Generation}

\title{A Hybrid Artificial Intelligence System for Automated EEG Background Analysis and Report Generation}
\author{Chin-Sung Tung, 
        Sheng-Fu Liang, \IEEEmembership{Member, IEEE}, 
        Shu-Feng Chang, 
        Chung-Ping Young, \IEEEmembership{Member, IEEE}
\thanks{This work did not receive any external financial support.  \emph{(Corresponding author: Sheng-Fu Liang; Chung-Ping Young.)}}
\thanks{Chin-Sung Tung is with the Department of Neurology, Min-Jhong Hospital, Pingtung 900, Taiwan 
and is aslo pursuing a Master's degree in Computer Science and Information Engineering, National Cheng Kung University, Tainan 70101, Taiwan (e-mail: p77121082@gs.ncku.edu.tw).}
\thanks{Sheng-Fu Liang and Chung-Ping Young are with the Department of Computer Science and Information Engineering, National Cheng
Kung University, Tainan 70101, Taiwan (e-mail: sfliang@mail.ncku.edu.tw; dryncku@gmail.com).}
\thanks{Shu-Feng Chang is with the Department of Neurology, Min-Jhong Hospital, Pingtung 900, Taiwan (e-mail: neurcsf@gmail.com).}
}


\maketitle

\begin{abstract}
Electroencephalography (EEG) plays a crucial role in the diagnosis of various neurological disorders. However, small hospitals and clinics often lack advanced EEG signal analysis systems and are prone to misinterpretation in manual EEG reading. This study proposes an innovative hybrid artificial intelligence (AI) system for automatic interpretation of EEG background activity and report generation. The system combines deep learning models for posterior dominant rhythm (PDR) prediction, unsupervised artifact removal, and expert-designed algorithms for abnormality detection. For PDR prediction, 1530 labeled EEGs were used, and the best ensemble model achieved a mean absolute error (MAE) of 0.237, a root mean square error (RMSE) of 0.359, an accuracy of 91.8\% within a 0.6 Hz error, and an accuracy of 99\% within a 1.2 Hz error. The AI system significantly outperformed neurologists in detecting generalized background slowing (p$=$0.02; F1: AI 0.93, neurologists 0.82) and demonstrated improved focal abnormality detection, although not statistically significant (p$=$0.79; F1: AI 0.71, neurologists 0.55). Validation on both an internal dataset and the Temple University Abnormal EEG Corpus showed consistent performance (F1: 0.884 and 0.835, respectively; p$=$0.66), demonstrating generalizability. The use of large language models (LLMs) for report generation demonstrated 100\% accuracy, verified by three other independent LLMs. This hybrid AI system provides an easily scalable and accurate solution for EEG interpretation in resource-limited settings, assisting neurologists in improving diagnostic accuracy and reducing misdiagnosis rates.
\end{abstract}


\section{Introduction}
\label{sec:introduction}
\IEEEPARstart{E}{lectroencephalography} (EEG) plays a crucial role in clinical neurology, aiding in the diagnosis of various neurological disorders such as epilepsy, dementia, brain lesions, and localized cerebral structural abnormalities \cite{RN1}, \cite{eeg_Significance}. With the advent of an aging society, the incidence of neurodegenerative diseases and cerebral structural disorders, including dementia, stroke, traumatic subdural hemorrhage, and brain tumors, is gradually increasing \cite{nuero_burden}. In small hospitals or neurological clinics, EEG serves as a first-line diagnostic tool due to its relative low cost, non-invasiveness, absence of radiation exposure, and repeatability. The interpretation of EEG background activity, which includes PDR, symmetry, and the presence of focal slow waves, is an essential method for assessing cerebral cortical function and facilitates the diagnosis of these disorders \cite{RN2}. In healthcare facilities lacking radiological equipment such as Computed Tomography or Magnetic Resonance Imaging, EEG becomes even more important. However, a significant proportion of hospitals or clinics  lack the capability for quantitative analysis and assisted interpretation of EEG features and reports often rely on the experience of clinicians. Low inter-rater agreement among neurologists interpreting clinical EEG is a significant problem and may lead to inconsistencies in diagnosis and treatment decisions \cite{JAMA_Neurology_2023}. 

The SCORE-AI system\cite{JAMA_Neurology_2023} uses an artificial intelligence (AI) approach for automatic interpretation of clinical EEG, providing a solution to these challenges in EEG interpretation. The model employs a convolutional neural network architecture trained on a large dataset of 30,493 highly annotated EEGs to classify recordings as normal or into clinically relevant abnormal categories. One limitation is that the SCORE-AI system requires a large amount of training data labeled by experts, which may not be feasible for small hospitals and clinics. Furthermore, the complexity of EEG labeling is relatively high, and the SCORE program may not be easily integrated into existing EEG and hospital information systems, making it less suitable for small healthcare institutions.

Additionally, the presence of numerous artifacts in clinical EEG and the inability to achieve the same quality as research-grade EEG pose significant challenges for EEG preprocessing. Currently, there is no single unsupervised and efficient method that can effectively address these issues\cite{remove_artifacts}. Furthermore, structured EEG features were difficult to obtain in small institutions because commercial systems generally do not provide access to such data. Considering these challenges, this study aims to develop an AI-based automatic reporting algorithm for EEG background, which we refer to as the Hybrid AI system, using a smaller dataset based on routine clinical EEG for training and focusing on EEG background analysis. The system employs a hybrid model that includes anomaly detection, deep learning, guidelines, and expert heuristics.

In clinical practice, a report needs to be written following EEG analysis. To the best of our knowledge, many experts still write EEG reports manually, either through customized phrase input in the hospital information system or by using word processing software to enter the report form, especially in smaller institutes. With the advancements in large language models, an increasing number of researchers have started adopting AI-assisted generation in medical fields\cite{largeAIHealthInformatics}, \cite{RN4}. To date, there is no existing method for AI-generated EEG report content. This study also pioneers the use of large language models and EEG features to assist in generating EEG report content.

\begin{figure*}[ht]
    \includegraphics[width=\linewidth]{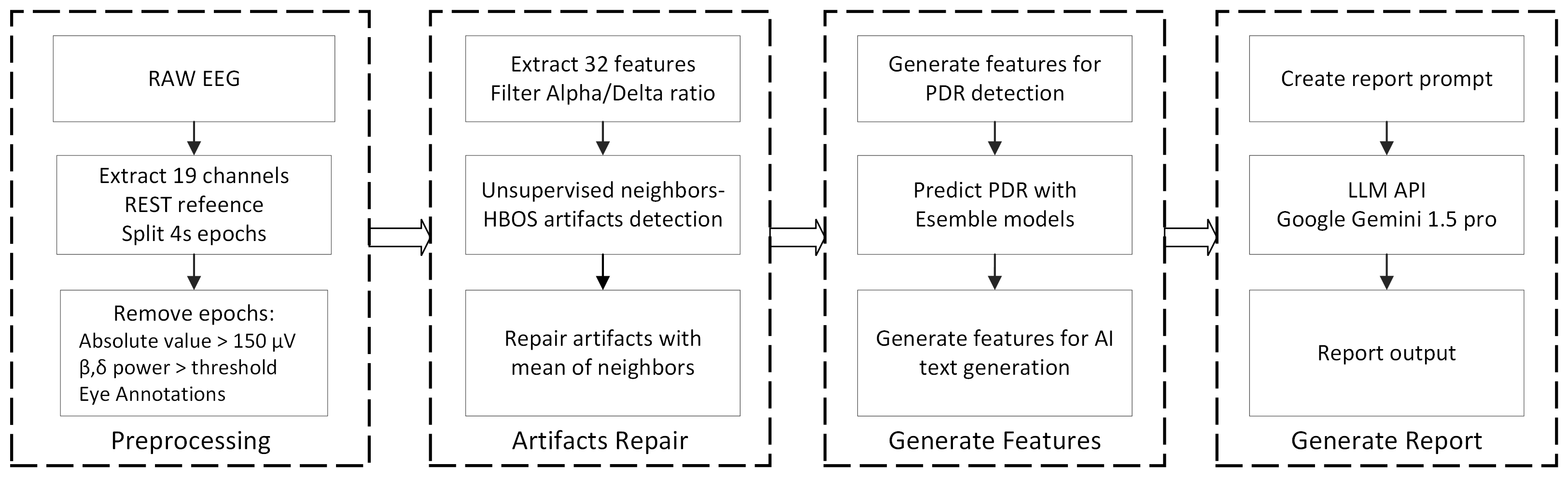}
    \caption{Workflow of the proposed hybrid AI system for automated EEG background interpretation and report generation.}
    \label{fig:fig_workflow}
\end{figure*}
\begin{figure*}[ht]
    \includegraphics[width=\linewidth]{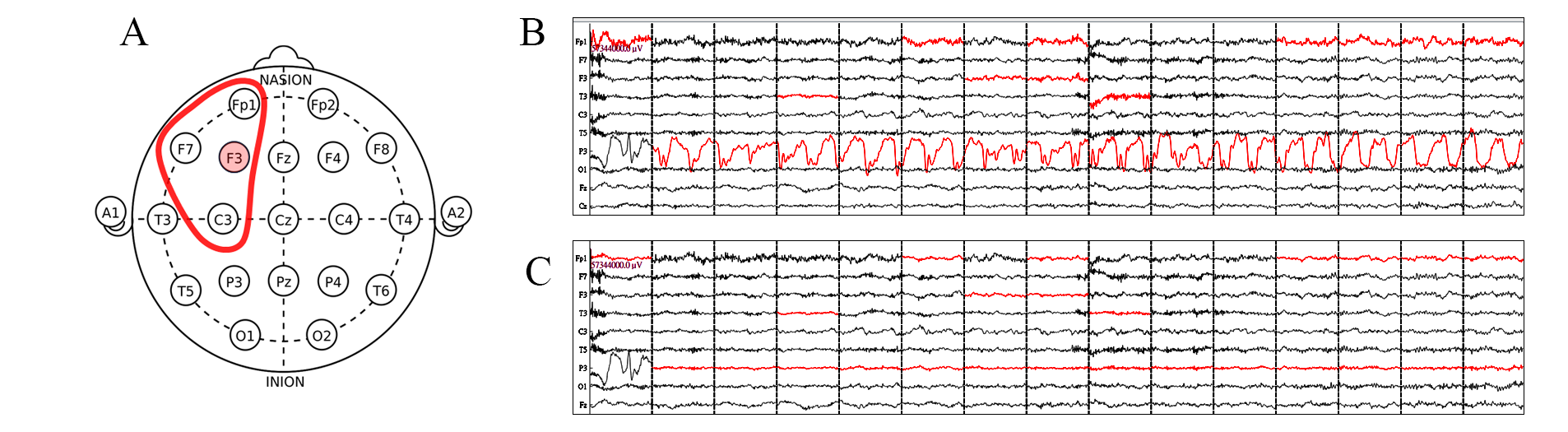}
    \caption{Neighbor electrode example and artifact repair using the neighbor-HBOS method. (A) F3 and its neighboring electrodes: Fp1, F7, and C3. (B) EEG signal before artifact repair. (C) EEG signal after artifact repair using the neighbor-HBOS method.}
    \label{fig:fig_neighbor_HBOS}

\end{figure*}

\section{Methods}
\label{sec:methods}

\subsection{Ethical Considerations}
This study was reviewed and approved by the Institutional Review Board (IRB) of National Cheng Kung University Hospital (NCKUH) (IRB No: B-ER-113-214). The requirement for informed consent was waived by the IRB, as the study involved the retrospective analysis of EEG data and medical records from March 1, 2020, to May 15, 2024, at Min-Jhong Hospital. The study was conducted in accordance with the Declaration of Helsinki and the ICH-GCP guidelines.

\subsection{EEG Data} 
\subsubsection{Acquisition} This study utilized a single-center retrospective clinical dataset, including all raw EEG files recorded at Pingtung Min-Jhong Hospital in Taiwan from March 2020 to May 2024. After excluding files without reports, unreadable files, or files that could not be opened, the total number of usable EEG recordings was 2,491.
\subsubsection{EEG Equipment and Recording} The study employed a Nicolet digital EEG machine for clinical use. Non-invasive methods were used to record EEG signals through standard scalp electrodes at 21 locations according to international 10-20 system. Each recording lasted approximately 10 minutes and included photic stimulation and hyperventilation procedures. The sampling rate was 125 Hz and the reference electrode was placed on the Fpz position. Fig. \ref{fig:fig_workflow} illustrates the entire system processing workflow.

\subsection{EEG Data Preprocessing}
\subsubsection{EEG File Conversion} Since our EEG format is not directly readable in the Python environment, MATLAB\cite{MATLAB} and the FieldTrip toolbox\cite{RN5} were used to convert Nicolet EEG raw files to the EDF format, and annotations were saved as plain text files. The primary EEG processing environment was Python 3.10, and the main EEG processing software was the open-source MNE-python\cite{RN6}.
\subsubsection{Rebuilding EEG Reference} There is no absolute standard for EEG reference electrode placement. The Reference Electrode Standardization Technique (REST)\cite{RN7}, \cite{RN8} approximates standardization to an infinite reference point and is a good choice for clinical EEG. In this study, the first step was to rebuild the reference electrode using the REST method and then segment the EEG into 4-second epochs.
\subsubsection{EEG Frequency Bands} Digital EEG is a type of multi-channel time-series data\cite{RN9}. EEG research began with the discovery of alpha waves, which range from 8 to 13 Hz and are most prominent during eyes-closed resting states, particularly in the posterior occipital region. This awake background alpha rhythm is referred to as the posterior dominant rhythm (PDR). Other clinically relevant EEG waves include beta waves (13 to 30 Hz), associated with active cognitive activity, anxiety, and drug use, and slow waves, including theta (4 to 8 Hz) and delta ($<$4 Hz). Slow waves can be present in EEGs of healthy individuals, but persistent slow waves are associated with cortical dysfunction. Widespread slow waves may be related to metabolic brain disorders, degenerative brain diseases, or extensive brain injury, while focal slow waves may indicate focal brain lesions or epileptiform abnormalities.
\subsubsection{Awake, Eyes-Closed EEG Segment Selection} Rigorous experiments use eyes-open and eyes-closed markers to determine awake, eyes-closed stages. However, in clinical settings, accurate labeling of awake resting-state EEG may be challenging due to factors such as patient non-compliance or omission of markers by busy technicians. Therefore, for automated analysis, we first calculated the band power of each EEG frequency to exclude segments that did not belong to the awake, eyes-closed state. Segments meeting the following criteria were removed: epochs that included annotations of eye open or close labels by the technicians during the examination, epochs with high original signal values ($>$150$\mu$V), and epochs with a high beta or delta band power ratio ($>$mean band power + 2.2 SD). The removed segments might have severe artifacts or a PDR that was too high or too low, indicating non-eyes-closed EEG. After removing these segments, the remaining EEG segments had higher alpha power.
\subsection{EEG Artifact Handling}
EEG artifacts, such as eye movements, electrode instability, subject's body movements, and external electromagnetic interference, can easily contaminate the EEG. To achieve fully automated EEG analysis, a method that can automatically handle artifacts without manual labeling is necessary. 

In this study, an unsupervised outlier anomaly detection method\cite{RN10} was adopted and improved. In the original unsupervised method, 58 EEG features were applied, and we use 31 EEG features extracted here from the segmented EEG. Then, using this features, the Histogram-based Outlier Score (HBOS)\cite{RN11} model was applied, along with our custom-designed neighboring electrode comparison method shown in fig. \ref{fig:fig_neighbor_HBOS}, to detect artifact-contaminated electrodes. If the artifact originates from a single electrode, the artifact signal is less likely to propagate to adjacent electrodes. Therefore, using the HBOS method, artifact-contaminated electrodes can be detected. This neighboring electrode comparison method, as opposed to the previous and subsequent epoch comparison method, can directly detect single-electrode artifacts. If an electrode is determined to be artifact-contaminated, the signals from neighboring electrodes are averaged to repair the affected portion. 

Additionally, if the proportion of artifact-contaminated epochs for a single electrode exceeds 30\%, the electrode is listed as an artifact channel. During the experimental process, it was observed that alpha waves have the highest power in the occipital region and this may lead to false positives being considered as anomalies. Therefore, before detection, epochs with alpha power exceeding a threshold are excluded from anomaly detection. The threshold value is determined based on the mean and median values of the data, and empirically, values between these two values yield better results.

\subsection{Obtaining Features for Interpreting EEG Background}

\begin{figure}
    \centering
    \includegraphics[width=\linewidth]{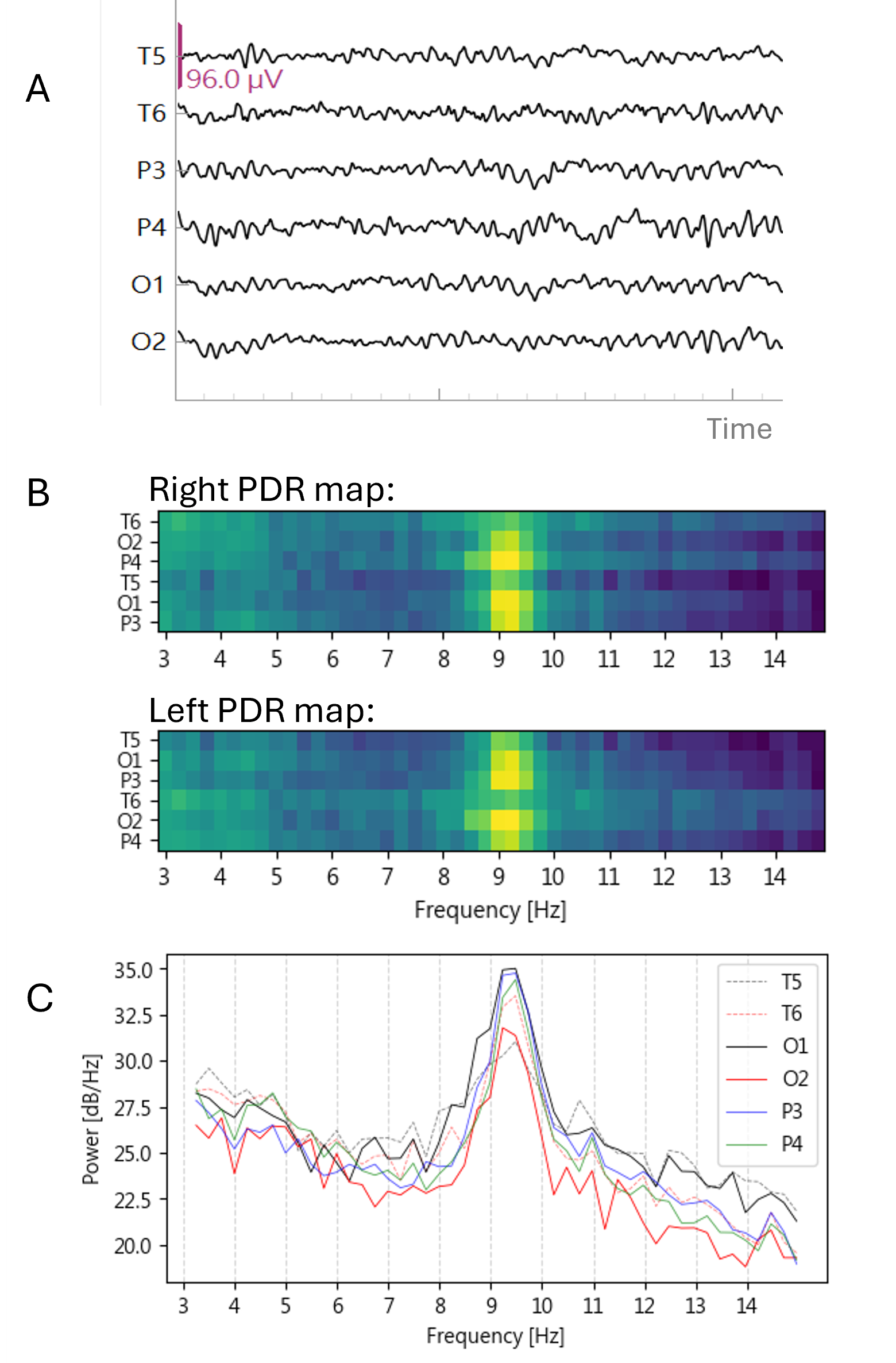}
    \caption{Example of features used for PDR prediction. (A) Original EEG with a labeled PDR value of 9.5 Hz for both the right and left hemispheres. The ensemble model's predicted values are 9.4 Hz for the right hemisphere and 9.5 Hz for the left hemisphere. (B) Feature maps for the right and left PDR, displayed as images. The right PDR map includes electrodes [T6, O2, P4, T5, O1, P3], while the left PDR map includes electrodes [T5, O1, P3, T6, O2, P4]. (C) Power spectral density (PSD) features for each electrode, showing the power distribution across frequencies from 3 to 15 Hz.}
    
    \label{fig:fig_PDRfeature}
\end{figure}
\subsubsection{Posterior Dominant Rhythm (PDR) Prediction:}
\begin{figure}
    \centering
    \includegraphics[width=\linewidth]{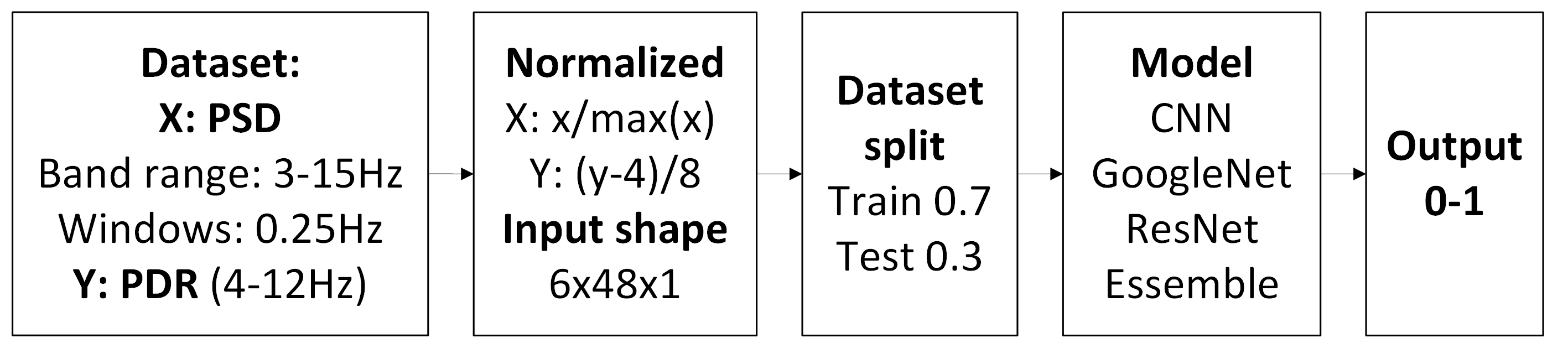}
    \caption{Overview of the deep learning pipeline for PDR prediction model. The input dataset X contains power spectral density (PSD) values with a frequency band range of 3 to 15Hz, window size of 0.25Hz, PDR range of 4 to 12Hz. The dataset is normalized with X values scaled by the maximum X value and Y values scaled by (y-4)/8. The normalized dataset is split into a training set (70\%) and test set (30\%). The training data is used to train a CNN model, which could be GoogleNet, ResNet, or an ensemble approach. The trained model outputs predictions between 0 and 1.}
    \label{fig:fig_models}
\end{figure}

    \paragraph{Dataset} 1,547 EEG data with expert-labeled background dominant rhythm were collected in this paper. After excluding 17 EEGs that were difficult to determine the dominant rhythm, 1,530 data remained. The data were divided into training and testing sets using artifact-repair, artifact-non-repair, and mixed data, with 70\% for training and 30\% for testing (Fig. \ref{fig:fig_models}). The file name was used as the clustering index, and the data were evenly distributed according to the label value to ensure balanced data clustering and prevent duplication in the training or testing sets. The PDR of the left and right EEG were labeled separately. With this approach, the mixed dataset could be expanded to 6,120 data. 
    
    \paragraph{Labeling} PDR prediction was realized by training a supervised deep learning model. Experienced neurologists labeled the background PDR at 0.5 Hz intervals, with a range of 4 to 12 Hz, which covers the PDR range for most adults \cite{PDR_Automated_EEG_Analysis_2011}. All 1,530 included EEG recordings were labeled with the dominant rhythm.

    \paragraph{Features} The features used were the band power spectrum of the EEG, obtained from 6 posterior electrodes bilaterally (T6, O2, P4, T5, P3, O1). The sampling frequency range was 3 to 15 Hz, using the multitaper method with a sampling interval of 0.25 Hz, resulting in a 6x48x1 feature matrix (Fig. \ref{fig:fig_PDRfeature}). The multitaper method was chosen for its advantages of providing high frequency resolution and low variance estimates of the power spectral density\cite{Multitaper_1982}, \cite{Multitaper_2014}. For data predicting the right EEG PDR, the right electrodes were placed in the first 3 rows of the matrix (T6, O2, P4, T5, O1, P3), and for predicting the left EEG PDR, the order was reversed (T5, O1, P3, T6, O2, P4). The features were standardized to a range of 0 to 1. These features were used as X, and the frequency values of the labels were used as Y for deep learning training.

    \paragraph{Models} Recent studies utilizing deep learning models, especially convolutional neural networks (CNNs), have demonstrated success in multi-task classification for EEG applications such as time-frequency analysis\cite{eegDeepLearning}, prediction of stroke patient recovery\cite{deepLearningStroke}, epileptic EEG diagnosis\cite{deepLearningEpilepsy}, and motor brain-computer interfaces\cite{deepLearningBCI}. Therefore, CNN models were chosen for PDR prediction. Due to the small feature dimension, three relatively low-weight neural network architectures were used, including a custom-defined CNN\cite{RN13}, GoogleNet\cite{RN14}, ResNet\cite{RN15}, and an ensemble model architecture. Both GoogleNet and ResNet were used without pre-trained weights. Regression-based CNN models were chosen for PDR prediction, as they allow for more precise frequency predictions compared to classification, given the continuous nature of EEG frequencies. The labels were normalized from their original 0.5 Hz interval range of [4, 12] Hz to [0, 1] to match the sigmoid output activation function used in the final layer of each model. All models use the Mean Squared Error (MSE) loss function and the Adam optimizer. Table~\ref{tab:model_structures} provides a detailed overview of each model's architecture.
    \begin{table}
        \caption{Model Architectures}
        \begin{tabular}{p{1.6cm}p{6.4cm}}
        \hline
        \textbf{Model} & \textbf{Key Architecture Components} \\
        \hline
        Custom CNN & \begin{itemize}
        \item Input shape: (6, 48, 1)
        \item 5 Conv2D layers: 64, 128, 256, 512, 512 filters
        \item 3 MaxPooling2D layers
        \item Dense: 64 units, ReLU activation
        \item Dropout: 0.2 rate
        \item Output: 1 unit, sigmoid activation
        \end{itemize} \\
        \hline
        GoogleNet & \begin{itemize}
        \item Input shape: (6, 48, 1)
        \item Initial Conv2D: 64 filters, MaxPooling2D
        \item 5 Inception modules
        \item GlobalAveragePooling2D
        \item Dense: 128 units, ReLU activation
        \item Output: 1 unit, sigmoid activation
        \end{itemize} \\
        \hline
        ResNet & \begin{itemize}
        \item Input shape: (6, 48, 1)
        \item Initial: 2 Conv2D layers (32, 64 filters)
        \item 3 Residual blocks: 128, 256, 512 filters
        \item Each block: Shortcut + Main path
        \item GlobalAveragePooling2D
        \item Output: 1 unit, sigmoid activation
        \end{itemize} \\
        \hline
        \multicolumn{2}{l}{Common features: Loss - Mean Squared Error, Optimizer - Adam} \\
        
        \end{tabular}
        
        \label{tab:model_structures}
        \end{table}

    \paragraph{Training} The training environment utilized an Nvidia RTX 4090 GPU, Intel(R) i7--13700 2.10 GHz CPU, 64GB RAM, and the Windows 11 operating system. The programming environment was Python 3.10 with Keras 2.10\cite{chollet2015keras}. The training parameters were set to 200 epochs and a batch size of 16. The activation function of the output layer was sigmoid. Each model was trained 10 times, and all training results were recorded. Finally, the best MAE result for each model was selected to form an ensemble model for comparison.

    \paragraph{Validation Methods}
    Two approaches were used to assess model performance, both were tested with single GoogleNet model and repair-artifact data:
    \begin{itemize}
    \item K-fold Cross-validation: The dataset was split into 4 folds, with 3 for training and 1 for testing. This process was repeated 10 times, rotating the test set.
    \item Smaller Dataset Validation: Models were trained on different dataset ratios (0.2, 0.4, 0.6, 0.8) of the original size, maintaining a 7:3 train-test split. Each ratio was randomly sampled in the training and testing set and run 10 times to determine the minimum data required for accurate PDR prediction.
    \end{itemize}
    \paragraph{Performance Metrics}
    Model performance was evaluated using RMSE, MAE, R2, and accuracy. Accuracy was further divided into ACC0.6 and ACC1.2, representing the proportion of predictions with errors $<$0.6 Hz and $<$1.2 Hz, respectively \cite{PDR_Automated_EEG_Analysis_2011}.
    \paragraph{Statistical Analysis}
    Statistical tests were conducted to compare performance across models, datasets, and validation methods:
    \begin{itemize}
    \item Main results: Paired t-tests for RMSE and MAE; McNemar's test for ACC0.6 and ACC1.2.
    \item K-fold validation: Independent t-tests for MAE and RMSE; chi-square tests for ACC0.6 and ACC1.2.
    \item Smaller dataset validation: ANOVA followed by Tukey's HSD test to compare all training results across dataset ratios.
    \end{itemize}

\subsubsection{Obtaining Other EEG Features} Using the multitaper method with a sampling interval of 0.25 Hz, the band power spectrum was obtained, and features that could be used to interpret the EEG background were calculated, including:
    \paragraph{Anterior-posterior gradients (AP gradient)\cite{RN16}} The anterior electrodes included Fp1, Fp2, F7, F8, F3, and F4, while the posterior electrodes included T5, T6, P3, P4, O1, and O2. The band power was calculated separately for the left, right, and total electrodes. The AP gradient was calculated by dividing the total alpha band power of the anterior electrodes by the total alpha band power of the posterior electrodes. An AP gradient below 40\% was considered normal\cite{RN16}.
    $$AP_{gradient} = \frac{\sum_{i \in A} P_{\alpha, i}}{\sum_{j \in P} P_{\alpha, j}}$$
    where $A$ is the set of anterior electrodes and $P$ is the set of posterior electrodes. 
    \paragraph{Total power} The total band power between 1.5 to 30 Hz was calculated separately for the left, right, and total electrodes, resulting in three total power values.
    $$P_{total, k} = \int_{1.5}^{30} P_k(f) df$$
    where $k$ is the electrode index.
    \paragraph{Slow band power ratio (1.5 to 8 Hz)\cite{RN16}} The band power between 1.5 to 8 Hz was calculated separately for the left, right, and total electrodes, and then divided by the corresponding total power. A ratio below 60\% was considered normal\cite{RN16}.
    $$R_{slow, k} = \frac{\int_{1.5}^{8} P_k(f) df}{P_{total, k}} \times 100\%$$
    where $k$ is the electrode index.
    \paragraph{Left-right alpha, theta, and delta band power ratios} To calculate the left-right power ratios for the alpha, beta, and delta bands, the following formula can be used:
    $$R_{left/right, band} = 2 \times \frac{P_{left, band} - P_{right, band}}{P_{left, band} + P_{right, band}}$$
    {where $band$ is the frequency band: alpha (8 to 13 Hz), theta (4 to 8 Hz), or delta (1.5 to 4 Hz).}
    
    Since the interpretation of left-right symmetry and focal abnormalities is more complex and easily affected by artifacts, these values cannot be directly used to determine abnormalities. A scoring algorithm, described below, will be used to make this determination.

\subsection{Algorithm for Interpreting EEG Abnormalities}
The EEG background abnormality algorithm, based on guidelines\cite{RN17} and quantified techniques\cite{RN16}, relies on carefully determined thresholds for effective anomaly analysis. Table~\ref{tab:anomaly_thresholds} outlines key parameters, thresholds, and determination methods, derived from expert knowledge, literature, and dataset analysis. The algorithm consists of three main components:

\begin{table}
    \renewcommand{\arraystretch}{1.3}
    \caption{Threshold Determination Methods for Anomaly Analysis Algorithm}
    \centering
    \begin{tabular}{lcp{3.5cm}}
    \hline
    \textbf{Parameter} & \textbf{Threshold} & \textbf{Determination Method} \\
    \hline
    PDR & $<$ 7.5 Hz & Consensus among neurologists \\
    \hline
    Slow band power ratio & $<$ 50\% & \multirow{3}{3.5cm}{Based on previous literature\cite{RN16} \cite{RN17}} \\
    L-R band power ratio* & $>$ 50\% & \\
    L-R PDR difference & $>$ 1 Hz &  \\
    \hline
    Focal abnormality score & $>$ 2.4 & \multirow{2}{3.5cm}{5\% of data distribution} \\
    Alpha amplitude score & $>$ 1.6 &  \\
    \hline
    \multicolumn{3}{l}{L-R: Left-right, PDR: Posterior dominant rhythm} \\
    \multicolumn{3}{l}{*Includes alpha, theta, and delta bands} \\
    \end{tabular}
    
    \label{tab:anomaly_thresholds}
    \end{table}

\subsubsection{Generalized Background Slowing (GBS)} GBS is an important EEG abnormality that indicates global cerebral dysfunction\cite{EEG_Abnormal}. GBS is characterized by the presence of slowing in the theta (4 to 8 Hz) or delta (1.5 to 4 Hz) frequency ranges, which can be of high or low amplitude. In our algorithm, GBS is considered present if any of the following criteria are met:
\begin{itemize}
    \item Right PDR and left PDR $<$7.5 Hz
    \item Right PDR and left PDR $<$8 Hz and Slow band power ratio $>$ 50\%
\end{itemize}
\subsubsection{Background Asymmetry} Background asymmetry is divided into two parts: a frequency difference that is too large (left-right PDR difference $>$1Hz) or an amplitude difference greater than 50 \%. The following method is used to determine background asymmetry, and any of the conditions is considered asymmetric.
\begin{itemize}
    \item PDR difference $>$ 1 Hz
    \item Alpha amplitude score $>$ 1.6 (\hyperref[alg:alpha_amplitude]{Algorithm 1}): The left-right alpha band power ratio is used as an indicator, taking 14 electrodes F8, F4, C4, T4, T6, P4, O2, F7, F3, C3, T3, T5, P3, and O1, and excluding electrodes identified as artifacts. The score is calculated for the left and right sides. If the absolute alpha band ratio $>$ 0.5, the absolute ratio is added to the corresponding side. If the score is greater than 1.6, the background is considered asymmetric.
    
\end{itemize}
\begin{algorithm}
    \label{alg:alpha_amplitude}
    \caption{Alpha Amplitude Score Algorithm}
    \begin{algorithmic}
    \REQUIRE $R_{\alpha}$ \COMMENT{Left-right alpha band power ratios}
    \REQUIRE $artifacts$ \COMMENT{Set of artifact electrodes}
    \REQUIRE $electrodes$ \COMMENT{Set of all electrodes}
    \ENSURE $score_{left}$, $score_{right}$ \COMMENT{Left and right alpha amplitude scores}
    \ \textit{Initialization} :
    \STATE $score_{left} \gets 0$
    \STATE $score_{right} \gets 0$
    \FOR{each electrode $i \in electrodes$}
    \IF{$i \notin artifacts$}
    \IF{$|R_{\alpha}[i]| > 0.5$}
    \IF{$R_{\alpha}[i] > 0$}
    \STATE $score_{left} \gets score_{left} + |R_{\alpha}[i]|$
    \ELSE
    \STATE $score_{right} \gets score_{right} + |R_{\alpha}[i]|$
    \ENDIF
    \ENDIF
    \ENDIF
    \ENDFOR
    \IF{$score_{left} > 1.6$ \OR $score_{right} > 1.6$}
    \RETURN Asymmetric background amplitude
    \ELSE
    \RETURN Symmetric background amplitude
    \ENDIF
    \end{algorithmic}
\end{algorithm}
\begin{algorithm}
    \caption{Focal Slow Waves Algorithm}
    \begin{algorithmic}
    \REQUIRE $R$ \COMMENT{Left-right theta and delta band power ratios, with $R[i, 0]$ representing theta and $R[i, 1]$ representing delta for electrode $i$}
    \REQUIRE $artifacts$ \COMMENT{Set of artifact electrodes}
    \ENSURE $score_{left}$, $score_{right}$ \COMMENT{Left and right focal abnormality scores}
    \ENSURE $E_{abn}$ \COMMENT{Set of abnormal electrodes}
    \ \textit{Initialization} :
    \STATE $score_{left} \gets 0$
    \STATE $score_{right} \gets 0$
    \STATE $E_{abn} \gets \{\}$
    \STATE \textbf{function} $neighbors(i)$
    \STATE \hspace{\algorithmicindent} \textbf{return} adjacent electrodes of $i$
    \STATE \textbf{end function}
    \FOR{each electrode $i$}
    \IF{$i \notin artifacts$}
    \IF{$(|R[i, 0]| > 0.5$ \OR $|R[i, 1]| > 0.5)$ \AND $\exists j \in neighbors(i) : (|R[j, 0]| > 0.5$ \OR $|R[j, 1]| > 0.5)$}
    \STATE $E_{abn} \gets E_{abn} \cup \{i\}$
    \ENDIF
    \ENDIF
    \ENDFOR
    \FOR{each electrode $i \in E_{abn}$}
    \FOR{$k \in \{0, 1\}$}
    \IF{$|R[i, k]| > 0.5$}
    \IF{$R[i, k] > 0$}
    \STATE $score_{left} \gets score_{left} + |R[i, k]|$
    \ELSE
    \STATE $score_{right} \gets score_{right} + |R[i, k]|$
    \ENDIF
    \ENDIF
    \ENDFOR
    \ENDFOR
    \IF{$score_{left} > 2.4$ \OR $score_{right} > 2.4$}
    \RETURN Focal slow waves present, $E_{abn}$
    \ELSE
    \RETURN No focal slow waves
    \ENDIF
    \label{alg:focal_slow}
    \end{algorithmic}
\end{algorithm}
\subsubsection{Focal Slow Waves} Focal slowing is another important EEG abnormality that indicates focal cerebral dysfunction. It can be continuous or intermittent and is characterized by the presence of slow waves in a specific brain region\cite{EEG_Abnormal}.

For interpreting focal slow waves, the electrodes at the same location on the left and right are used as references. The left-right theta and delta band power ratios are used as indicators. Based on empirical rules, we designed a algorithm (\hyperref[alg:focal_slow]{Algorithm 2}) that also uses the neighboring electrode comparison method to determine the presence of focal abnormalities.
\begin{itemize}
    \item The left-right theta and delta band power ratios are used as indicators. Since the numerator is left-side power minus right-side power, a positive value indicates that the left-side band power is larger, while a negative value indicates that the right-side band power is larger.
    \item If the theta or delta ratio of a single electrode is $>$ 0.5, and at least one of its neighboring electrodes also has a ratio $>$ 0.5 for either measure, the electrode is considered abnormal. Artifact electrodes are not included in the calculation.
    \item After obtaining all abnormal electrodes, the abnormality score is calculated separately for the left and right sides. If the absolute theta or delta band power ratio of an electrode is $>$ 0.5, the absolute value of the corresponding ratio is added to the score of the respective side
    \item Finally, if the unilateral focal abnormality score is greater than 2.4, focal slow waves are considered present in the brain region corresponding to the abnormal electrodes.
    
\end{itemize}

\subsection{Validating the Accuracy of the Hybrid AI System}
\subsubsection{Data Sources and Methodology}
Two datasets were used for validation:

\paragraph{Custom Validation Dataset}
The validation dataset consisted of 100 EEG recordings that were not used in the PDR model development. The ground truth labels were established by majority agreement among three neurologists who independently reviewed the EEGs. The two categories of abnormalities labeled were: 1) generalized background slowing (GBS), and 2) focal abnormalities, which included background asymmetry or focal slow waves. The neurologists were all blinded to the patients' information, original reports, and the results of the hybrid AI system.
The original EEG reports, as interpreted by neurologists were retrieved. Each report was manually labeled to indicate whether its content mentioned the two aforementioned abnormality indicators.
The AI hybrid system was then used to analyze the EEGs to output the inicaters of GBS and focal abnormalities. The system's output was compared to the neurologists' labels and the original reports to assess its accuracy.

\paragraph{Public TUAB Dataset}
276 EEG recordings from evaluation set of Temple University Abnormal EEG Corpus v3.0.1 \cite{Lopez2017TUH}. The initial dataset consisted of 150 normal and 126 abnormal EEGs. After a rigorous review process involving three neurologists, who unanimously agreed on the reclassification, 28 EEGs were relabeled to ensure consistency with our study's focus on background abnormalities. This expert review and reclassification resulted in a final dataset comprising 178 normal and 98 abnormal EEGs. The abnormal group is corresponded to any of the GBS or focal abnormalities.
The hybrid AI system was used to analyze the TUAB EEGs, and the output was compared to the expert-labeled ground truth to evaluate the system's performance.

\paragraph{TUAB Dataset Preprocessing}
The TUAB recordings were cropped to the first 600 seconds for several key reasons:
\begin{itemize}
    \item To maintain consistency with our custom dataset's average duration (548.80 ± 44.16 seconds vs. TUAB's original 1372.40 ± 594.60 seconds).
    \item To focus on the most diagnostically relevant portion of the EEG. Recordings after 600 seconds often showed signs of subject sleepiness, resulting in a slower background that could potentially confound the analysis of non-epileptic background abnormalities.
    \item Preliminary analysis of the PDR of the TUAB dataset showed a significant decrease after 600 seconds (8.64 ± 1.80 Hz to 8.45 ± 1.75 Hz, p $<$ 0.001), suggesting the onset of drowsiness and supporting our decision to crop recordings at 600 seconds.
\end{itemize}

\subsubsection{Performance Metrics and Statistical Analysis}
\paragraph{Custom Dataset}
Gwet's AC1 coefficient was used to assess inter-rater agreement for the neurologists' ground truth labels. Metrics for comparing the AI system's performance to the neurologists' original reports included confusion matrices, F1 score, accuracy, precision, recall. McNemar's test was used for statistical significance testing of the performance differences between the AI system and the neurologists.

\paragraph{TUAB Dataset}
EEG features across normal, abnormal, and relabeled groups analyzed using mean and standard deviation. Statistical significance was determined using two-tailed t-tests. The performance metrics used for the TUAB dataset were the same as those for the custom dataset. Z-tests were used to compare the AI system's performance between the custom dataset and the TUAB dataset.

\begin{figure}
    \centering
    \includegraphics[width=\linewidth]{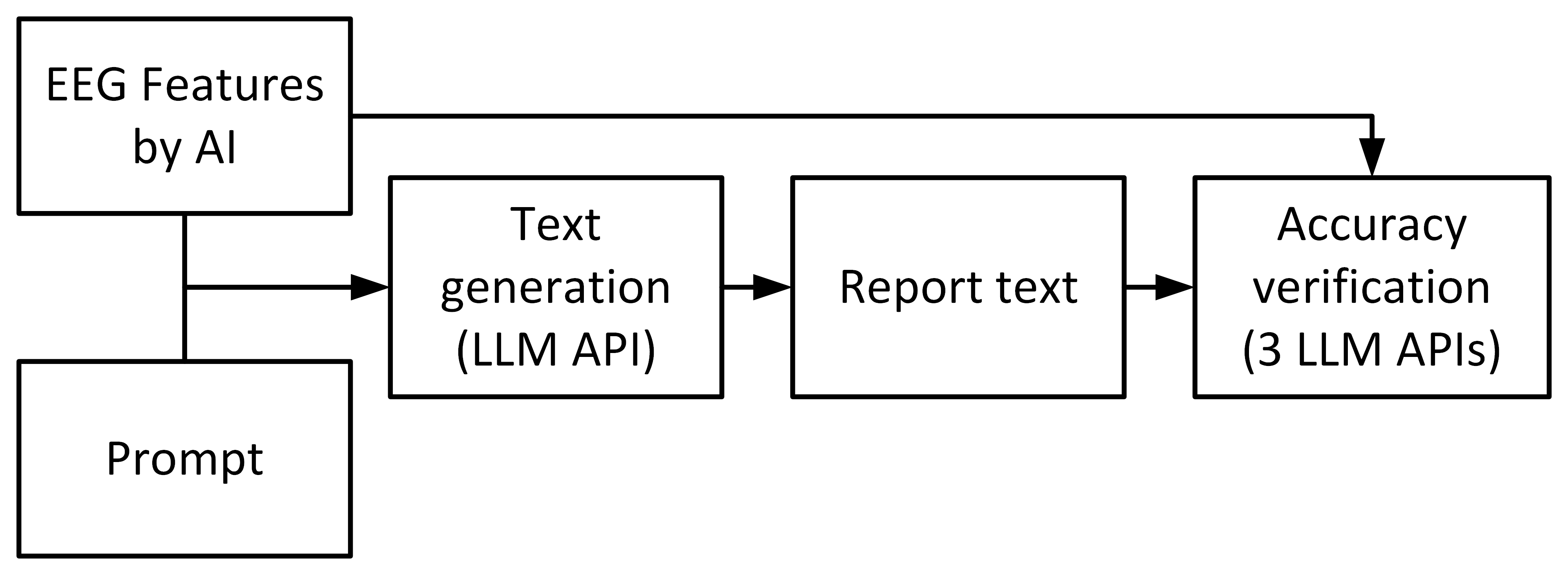}
    \caption{The workflow for generating and verifying EEG reports using large language models (LLMs). The process begins with an input prompt that combines the EEG features. This prompt is then passed to the Google Gemini 1.5 Pro API for text generation. The API generates a report text, which subsequently undergoes an accuracy verification step using an ensemble of three LLMs: Gemini 1.5 Flash, Claude 3 Sonnet, and GPT-4o. These models independently assess the report's accuracy, and the final verification result is determined based on the majority agreement among the three LLMs.}
    \label{fig:fig_gen_and_verify}
\end{figure}
\subsection{Generating EEG Background Reports}
Figure \ref{fig:fig_gen_and_verify} illustrates the workflow for generating and verifying EEG reports.
The system uses a large language model (LLM), specifically the Google Gemini 1.5 Pro API\cite{geminiteam2024gemini} for its state-of-the-art performance across a wide range of tasks, including long-context retrieval, reasoning, and real-world applications, while maintaining high compute efficiency. 

\subsubsection{Input Processing}
The system takes as input the structured EEG features generated by our hybrid AI algorithm. These features are formatted as a JSON object, containing key metrics such as background frequency, amplitude, symmetry, and any detected abnormalities.

\subsubsection{Prompt Engineering}
A carefully designed prompt is crucial for guiding the LLM to generate accurate and clinically relevant reports. Our prompt structure includes:

\begin{itemize}
    \item \textbf{Role Definition:} Establishes the LLM as a neurologist with access to comprehensive neurological databases.
    \item \textbf{Data Provision:} Incorporates the structured EEG features and their interpretations.
    \item \textbf{Task Specification:} Directs the LLM to generate a detailed, structured EEG report.
    \item \textbf{Report Structure:} Outlines the required sections (Findings, Conclusion, Clinical Correlation, Advanced Strategies).
    \item \textbf{Guidance:} Provides examples and instructions for handling normal and abnormal findings.
\end{itemize}

\subsubsection{Report Generation}
The system sends the constructed prompt to the Google Gemini API, which then generates the EEG report. The resulting report adheres to the specified structure instructed by the prompt. To illustrate the process, we present a real example of JSON features and the corresponding generated report:

\paragraph{JSON Features}

\begingroup
    \fontsize{8.5pt}{9pt}\selectfont
\begin{verbatim} 
{ 'EEG_quality': 'Good',
  'bad_channels': ['Fp1', 'P3', 'F8', 'Pz'],
  'backgroundFrequency': 
    'Right: 8.6 Hz, Left: 8.3 Hz',
  'bg_active': 'Normal background frequency',
  'bg_amp': 'medium (10-50 mV)',
  'bg_amp_sym': 'lower in right',
  'bg_freq': 'symmetric',
  'abnormalFindings': [
    'Focal slow wave or asymmetric abnormality 
    detected;Lower alpha amplitude in F8, F4 
    channels']
}
\end{verbatim}
\endgroup

\paragraph{LLM Generated Report}
\begingroup
    \fontsize{8.5pt}{9pt}\selectfont
\begin{verbatim}
=== EEG Findings ===
- The EEG recording demonstrates good quality.
- Background activity: Normal frequency and 
  medium amplitude (10-50 mV). 
- Background frequency: 8.6 Hz (right), 8.3 Hz 
  (left), symmetric.
- Background amplitude: Lower in the right 
  hemisphere.
- Focal slowing and/or asymmetric abnormality: 
  Lower alpha amplitude in right frontal-temporal 
  region (F8, F4 channels).
=== Conclusion ===
Abnormal EEG findings are observed, characterized 
by focal slow wave or asymmetric abnormality in 
the right frontotemporal region.
=== Clinical Correlation ===
Focal slowing on EEG can be associated with an 
underlying structural brain abnormality in that 
region. 
=== Advanced Strategies ===
- Correlation with the patient's clinical 
  presentation and neurological examination is 
  necessary.
- Neuroimaging studies, such as MRI or CT scan of 
  the brain, are recommended to evaluate for 
  structural abnormalities.
\end{verbatim}
\endgroup

\subsection{LLMs Verification of Report Accuracy}
The accuracy of 512 EEG reports generated by the hybrid AI algorithm and LLM were verified using few-shot prompting with three LLM models: OpenAI GPT-4o mini\cite{gpt4o_2024_introducing}, Google Gemini 1.5 Flash \cite{geminiteam2024gemini}, and Claude 3.5 Sonnet \cite{anthropic_2024_introducing}. Few-shot prompting was performed on the report content to verify accuracy. The prompt instructed the LLM to classify the report content as 1 or 0 based on the presence of two indicators: GBS and focal abnormality. The LLM was instructed to return the values in an array format. The verification result was determined by the majority agreement of the three LLM models. The inter-rater agreement among the three LLMs was evaluated using Gwet's inter-rater reliability coefficients (AC1)\cite{Gwet_2008}. All 512 LLM verification results were compared with the results output by the hybrid AI algorithm on the two indicators of GBS and focal abnormality, and a confusion matrix was calculated to verify accuracy using the F1 score, accuracy, precision, and recall metrics. Finally, to ensure the accuracy of the AI-generated reports, human experts verified all the 512 reports generated by the hybrid AI system.

\section{Results}\label{sec:results}
\subsection{EEG Preprocessing Results}
A total of 2,491 usable EEG recordings were included in the preprocessing. Experiments were conducted separately with artifact removal and without artifact removal procedures, obtaining two sets of features for report generation: one with artifact correction and one without artifact correction. The comparison of EEG features between the artifact-repair and non-repair groups is shown in Table \ref{tab:EEG_features}.

\paragraph{Feature Analysis Results of Our Full Dataset}
Most EEG features did not show significant differences between the artifact-repair and non-repair groups. Features that exhibited statistically significant differences (p$<$0.05) included lower anterior-posterior gradient (AP gradient) in the Repair group, higher left-right hemisphere theta ratio in the Repair group, and higher left-right hemisphere alpha ratio in the Repair group. The Repair group had lower amplitude values in the beta, theta, and delta frequency bands, while the difference in alpha band amplitude was marginally significant. These differences indicate that the artifact removal process may affect certain aspects of the EEG signal, particularly the anterior-posterior gradient, hemisphere ratios of theta and alpha, and amplitude values of different frequency bands.

\begin{table}
    \caption{Comparison of EEG features between artifact-repair and non-repair groups.}
    \centering
    \begin{tabular}{lccc}
        \hline
        Feature & Artifact repair & Non-repair & p \\
        \hline
        AP gradient & 39.52±9.56 & 40.96±9.58 & $<$0.05 \\
        Right AP gradient & 39.63±10.44 & 40.82±10.64 & $<$0.05 \\
        Left AP gradient & 39.95±10.79 & 41.75±10.71 & $<$0.05 \\
        Right slow ratio & 52.75±20.40 & 52.77±20.35 & 0.98 \\
        Left slow ratio & 51.94±20.62 & 52.93±20.31 & 0.09 \\
        Total slow ratio & 52.38±20.33 & 52.85±20.15 & 0.41 \\
        Left/Right $\delta$ ratio & -0.01±0.23 & -0.01±0.22 & 0.9 \\
        Left/Right $\theta$ ratio & 0.00±0.21 & -0.00±0.20 & $<$0.05 \\
        Left/Right $\alpha$ ratio & -0.01±0.21 & -0.02±0.19 & $<$0.05 \\
        $\alpha$ Amplitude & 29.40±12.27 & 30.08±12.41 & 0.05 \\
        $\beta$ Amplitude & 30.34±18.68 & 31.60±19.45 & $<$0.05 \\
        $\theta$ Amplitude & 31.52±13.90 & 32.81±14.29 & $<$0.05 \\
        $\delta$ Amplitude & 41.06±18.79 & 44.18±20.00 & $<$0.05 \\
        \hline
    \end{tabular}
    \label{tab:EEG_features}
\end{table}

\paragraph{Feature Analysis Results of TUAB dataset}
Analysis of EEG features in the TUAB dataset revealed significant differences in nine of the thirteen features between abnormal and both normal groups, while only one feature showed a significant difference between the original and relabeled normal groups. These findings underscore the robustness of certain EEG characteristics, particularly slow wave features, in distinguishing abnormal from normal non-epileptic background EEG patterns and highlight the importance of precise labeling in EEG analysis.
\begin{table}[htbp]
    \centering
    \caption{Comparison of EEG Features Across Normal, Abnormal, and Relabeled Groups of TUAB Dataset}
    \begin{tabular}{lccp{1.5cm}}
    \hline
    \textbf{Feature} & \textbf{Normal} & \textbf{Abnormal} & \textbf{Normal (Relabeled)} \\
    \hline
    AP gradient & 36.61$\pm$7.35 & 48.93$\pm$12.32* & 39.24$\pm$7.39 \\
    Right AP gradient & 36.65$\pm$8.66 & 49.98$\pm$14.61* & 39.26$\pm$7.28 \\
    Left AP gradient & 36.87$\pm$7.74 & 48.44$\pm$11.74* & 39.40$\pm$8.93 \\
    Right slow ratio & 35.79$\pm$14.67 & 63.25$\pm$19.45* & 41.60$\pm$8.67$^\dagger$ \\
    Left slow ratio & 36.42$\pm$13.84 & 64.19$\pm$19.16* & 41.30$\pm$9.04 \\
    Total slow ratio & 36.08$\pm$14.21 & 63.76$\pm$19.10* & 41.44$\pm$8.77 \\
    Left/Right delta ratio & 0.03$\pm$0.31 & 0.04$\pm$0.45 & 0.02$\pm$0.30 \\
    Left/Right theta ratio & 0.04$\pm$0.33 & 0.04$\pm$0.45 & 0.03$\pm$0.29 \\
    Left/Right alpha ratio & 0.00$\pm$0.34 & 0.01$\pm$0.44 & 0.03$\pm$0.30 \\
    Alpha Amplitude & 28.68$\pm$12.38 & 22.86$\pm$9.79* & 27.98$\pm$9.86 \\
    Beta Amplitude & 28.12$\pm$13.00 & 31.25$\pm$21.99 & 27.26$\pm$9.22 \\
    Theta Amplitude & 22.68$\pm$9.70 & 29.76$\pm$13.56* & 26.01$\pm$10.49 \\
    Delta Amplitude & 25.25$\pm$11.53 & 41.95$\pm$21.04* & 25.76$\pm$8.44 \\
    \hline
    \multicolumn{4}{p{230pt}}{* Significantly different from Normal group (p $<$ 0.05)} \\
    \multicolumn{4}{p{230pt}}{$\dagger$ Significantly different from Normal group after relabeling (p $<$ 0.05)} \\

    \end{tabular}
    
    \label{tab:eeg-features}
    \end{table}

\begin{figure}
    \centering
    \includegraphics[width=\linewidth]{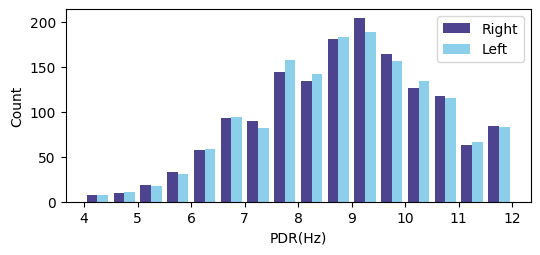}
    \caption{Distribution of the posterior dominant rhythm (PDR) in the dataset. The total count is 3060, with a mean of 8.63 Hz and a standard deviation (SD) of 1.66 Hz. The histogram shows the count of PDR values for the left and right hemispheres separately.}
    \label{fig:PDR_distribution}
\end{figure}

\subsection{PDR Prediction Results} Fig. \ref{fig:PDR_distribution} shows the distribution of the posterior dominant rhythm (PDR) in the dataset. The deep learning model used 1,530 EEG recordings with labeled PDR to train. The labeled PDR results had a mean of 8.6Hz, max of 12Hz, min of 4Hz, and a standard deviation of 1.66Hz.

\subsubsection{K-Fold Cross-Validation Results}
Table \ref{tab:kfold_results} presents the best MAE model results for each fold, along with the pairwise significance test results. The results show consistent performance across all folds, with MAE ranging from 0.255 to 0.281, ACC0.6 from 0.885 to 0.910, and ACC1.2 from 0.976 to 0.988. Pairwise significance tests revealed no statistically significant differences between folds for any evaluation metric (p \textgreater{} 0.05).
These findings demonstrate the models' ability to generalize well to unseen data. The consistent performance across different data subsets and the lack of statistically significant differences between folds further confirm the robustness and generalizability of the proposed PDR prediction models.
\begin{table}
\centering
\caption{K-fold cross-validation results}
\begin{tabular}{lccccc}
\hline
K Index & RMSE & MAE & r2 & ACC0.6 & ACC1.2 \\
\hline
K0 & 0.448 & 0.277 & 0.928 & 0.890 & 0.979 \\
K1 & 0.403 & 0.272 & 0.942 & 0.888 & 0.987 \\
K2 & 0.379 & 0.255 & 0.948 & 0.910 & 0.988 \\
K3 & 0.429 & 0.281 & 0.932 & 0.885 & 0.976 \\
\hline
\multicolumn{6}{p{230pt}}{p-values for comparison between folds:} \\
K0:K1 & 0.24 & 0.76 & - & 0.94 & 0.32 \\
K0:K2 & 0.07 & 0.18 & - & 0.24 & 0.23 \\
K0:K3 & 0.62 & 0.83 & - & 0.79 & 0.86 \\
K1:K2 & 0.38 & 0.26 & - & 0.18 & 1.00 \\
K1:K3 & 0.37 & 0.58 & - & 0.92 & 0.18 \\
K2:K3 & 0.07 & 0.10 & - & 0.13 & 0.12 \\
\hline
\multicolumn{6}{p{230pt}}{RMSE $=$ root mean square error, MAE $=$ mean absolute error, r2 $=$ R-squared value, ACC0.6 $=$ accuracy within 0.6 Hz, ACC1.2 $=$ accuracy within 1.2 Hz. P-values are from t-test for RMSE and MAE, and Chi-square test for ACC0.6 and ACC1.2.}
\end{tabular}
\label{tab:kfold_results}
\end{table}

\subsubsection{Smaller Dataset Validation Results}
\begin{figure}[!h]
\centering
\includegraphics[width=\columnwidth]{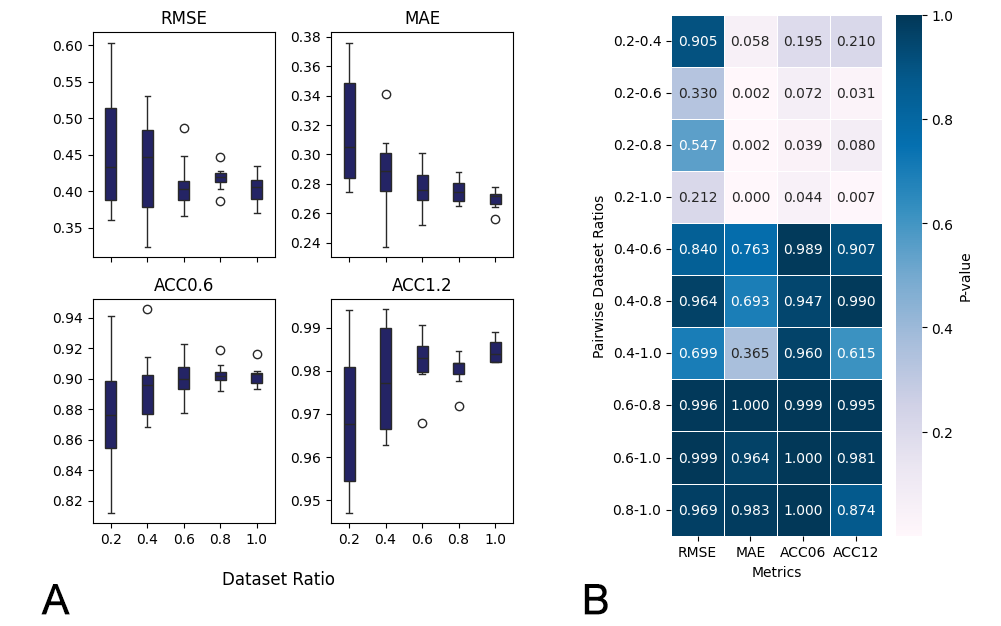}
\caption{Performance metrics for smaller dataset ratios: (A) RMSE, MAE, ACC0.6, and ACC1.2 values for each dataset ratio; (B) Pairwise comparison p-values between dataset ratios for each performance metric.}
\label{fig:smaller_dataset_results}
\end{figure}

Results from 10 runs of each dataset proportion are shown in Fig.~\ref{fig:smaller_dataset_results}. Performance generally improved with increasing dataset size, with the most significant improvements observed when increasing from 20\% to 40\% (RMSE $p=0.905$, MAE $p=0.058$, ACC0.6 $p=0.195$, ACC1.2 $p=0.210$). Further increases led to more gradual improvements, with non-significant differences between 60\%--80\% and 80\%--100\% ratios.
These findings suggest that 40\% of the original dataset (approximately 612 samples) may be sufficient for reasonably accurate PDR predictions. However, using 60\% to 80\% of the data (918--1,224 samples) provides better overall performance, with diminishing returns beyond 80\%.

\subsubsection{Performance on Different Datasets} (Table \ref{tab:pdr_prediction_results_datasets})
The mixed dataset (combining artifact-repair and non-repair EEG) achieved the best overall performance (RMSE: 0.365, MAE: 0.252, R-squared: 0.951, ACC0.6: 90.8\%, ACC1.2: 99.4\%). The non-repair dataset showed very similar results, with only slightly lower ACC0.6 (90.5\%) and ACC1.2 (99.2\%). The repair-only dataset demonstrated comparable performance (RMSE: 0.370, MAE: 0.256, R-squared: 0.949, ACC0.6: 91.6\%, ACC1.2: 98.7\%), notably achieving the highest ACC0.6.
Pairwise significance tests (t-test for RMSE and MAE, McNemar's test for ACC0.6 and ACC1.2) revealed no significant differences between the datasets for any metric (p \textgreater{} 0.05). This suggests that the model performs consistently across different EEG preprocessing approaches, demonstrating robustness in PDR prediction across various data conditions.
\begin{table}
\centering
\caption{Best results of PDR prediction for different training datasets}
\begin{tabular}{lccccc}
\hline
Dataset & RMSE & MAE & r2 & ACC0.6 & ACC1.2 \\
\hline
Mixed & 0.365 & 0.252 & 0.951 & 0.908 & 0.994 \\
Non-repair & 0.365 & 0.258 & 0.951 & 0.905 & 0.992 \\
Repair & 0.370 & 0.256 & 0.949 & 0.916 & 0.987 \\
\hline
\multicolumn{6}{p{230pt}}{p-values for comparison between datasets:} \\
Repair:Non-repair & 0.63 & 0.8 &  & 0.31 & 0.23 \\
Repair:Mixed & 0.69 & 0.56 &  & 0.46 & 0.096 \\
Non-repair:Mixed & 0.99 & 0.42 &  & 0.81 & 0.75 \\
\hline
\multicolumn{6}{p{230pt}}{RMSE $=$ root mean square error, MAE $=$ mean absolute error, r2 $=$ R-squared value, ACC0.6 $=$ accuracy within 0.6 Hz, ACC1.2 $=$ accuracy within 1.2 Hz. P-values are from t-test for RMSE and MAE, and McNemar's test for ACC0.6 and ACC1.2.}
\end{tabular}
\label{tab:pdr_prediction_results_datasets}
\end{table}

\subsubsection{Performance of Different Model Architectures} 
\begin{figure}
    \centering
    \includegraphics[width=\linewidth]{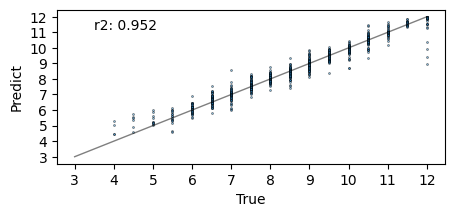}
    \caption{Scatter plot of the ensemble model's predicted values and true PDR values. The ensemble model achieves an R-squared value of 0.952, indicating a strong correlation between the predicted and true values.}
\label{fig_PDR_scatter}
\end{figure}

(Table \ref{tab:pdr_prediction_results_models})
The ensemble model achieved the best overall performance (RMSE: 0.359, MAE: 0.237, R-squared: 0.952, ACC0.6: 91.8\%, ACC1.2: 99.0\%), followed closely by GoogleNet (RMSE: 0.365, MAE: 0.252, R-squared: 0.951, ACC0.6: 90.8\%, ACC1.2: 99.4\%). CNN and ResNet performed relatively weaker but acceptably. Fig \ref{fig_PDR_scatter} shows the scatter plot of the ensemble model's predicted values and true PDR values.
Pairwise significance tests revealed that the ensemble model significantly outperformed CNN and ResNet in RMSE (p $<$ 0.001) and ACC1.2 (p $<$ 0.01). It also showed significantly lower MAE than CNN (p $<$ 0.001) and higher ACC0.6 than both CNN (p $=$ 0.04) and ResNet (p $=$ 0.02). GoogleNet significantly outperformed CNN in RMSE (p $=$ 0.07) and ACC1.2 (p $=$ 0.03), and ResNet in MAE (p $<$ 0.001) and ACC0.6 (p $=$ 0.91). No significant differences were found between GoogleNet and the ensemble model (p $>$ 0.05).
While artifact repair had limited impact, all models demonstrated considerable performance, validating the effectiveness of the proposed deep learning architectures for EEG PDR prediction.
\begin{table}
\centering
\caption{Best results of PDR prediction for different models}
\begin{tabular}{lccccc}
\hline
Model & RMSE & MAE & r2 & ACC0.6 & ACC1.2 \\
\hline
CNN & 0.428 & 0.276 & 0.932 & 0.900 & 0.986 \\
GoogleNet & 0.365 & 0.252 & 0.951 & 0.908 & 0.994 \\
ResNet & 0.420 & 0.251 & 0.935 & 0.898 & 0.978 \\
Ensemble & 0.359 & 0.237 & 0.952 & 0.918 & 0.990 \\
\hline
\multicolumn{6}{p{230pt}}{p-values for comparison between models:} \\
CNN:GoogleNet & 0.07 & 0.02 &  & 0.53 & 0.03 \\
CNN:ResNet & 0.62 & $<$0.01 &  & 0.91 & 0.15 \\
CNN:Ensemble & $<$0.01 & $<$0.01 &  & 0.04 & 0.29 \\
GoogleNet:ResNet & 0.07 & 0.91 &  & 0.41 & $<$0.01 \\
GoogleNet:Ensemble & 0.73 & 0.01 &  & 0.27 & 0.22 \\
ResNet:Ensemble & $<$0.01 & 0.01 &  & 0.02 & $<$0.01 \\
\hline
\multicolumn{6}{p{230pt}}{RMSE $=$ root mean square error, MAE $=$ mean absolute error, r2 $=$ R-squared value, ACC0.6 $=$ accuracy within 0.6 Hz, ACC1.2 $=$ accuracy within 1.2 Hz. P-values are from t-test for RMSE and MAE, and McNemar's test for ACC0.6 and ACC1.2.} \
\end{tabular}
\label{tab:pdr_prediction_results_models}
\end{table}

\subsection{Impact of Artifact Removal on EEG Dataset}
There are no reference indicators for artifact repair in the literature. Since the artifact removal uses unsupervised learning models, it is a challenging task to label the epochs with artifacts. In addition to presenting actual examples to observe the effect and correctness of artifact removal and signal restoration, another aspect is to see if there are differences in the PDR prediction results using the corrected dataset. Here, we assume that improper or excessive artifact correction will lead to poorer performance in PDR prediction. Another proof is to observe the accuracy of the AI algorithm's final abnormality prediction, which will be explained below.

\subsection{Accuracy Comparison between Hybrid AI System and Neurologist Interpretation}
The inter-rater agreement among the three neurologists was evaluated using Gwet's Agreement Coefficient (AC1). The results were as follows:
\begin{itemize}
\item For generalized background slowing (GBS), the AC1 was 0.61 (95\% CI: 0.49 to 0.73).
\item For focal abnormalities, the AC1 was 0.80 (95\% CI: 0.7 to 0.88).
\end{itemize}
These results indicate substantial agreement among the neurologists for focal abnormalities and moderate agreement for GBS, based on the interpretation of AC1 values \cite{Gwet_2008}.

Table \ref{tab:comparison_ai_neurologist} presents a comprehensive comparison between the Hybrid AI system and neurologist reports in detecting EEG abnormalities. For generalized background slowing (GBS), both AI models (with and without repair) significantly outperformed neurologists (p $=$ 0.02), achieving higher F1 scores (0.93 vs. 0.82), precision (0.95 vs. 0.74), and accuracy (0.94 vs. 0.83). However, neurologists showed slightly higher recall (0.93 vs. 0.90) for GBS. In focal slowing abnormality detection, the AI models demonstrated improvements, particularly in the repair mode. The AI-repair model achieved higher F1 score (0.71 vs. 0.55), recall (0.91 vs. 0.55), and accuracy (0.92 vs. 0.90) compared to neurologists, with a trade-off in precision (0.59 vs. 0.55). Notably, the AI-repair model showed substantial improvement over the non-repair version, especially in precision (0.59 vs. 0.48) and F1 score (0.71 vs. 0.62). However, the differences in focal slowing detection were not statistically significant (p $=$ 0.79). 

\begin{table}
\centering
\caption{Accuracy comparison of the Hybrid AI system and neurologist reports}
\begin{tabular}{lcccccc}
\hline
Reporter/Model & Category & F1 & P & R & ACC & P-value \\
\hline
Neurologist  & \multirow{3}{*}{GBS} & 0.82 & 0.74 & 0.93 & 0.83 &  \\
AI-non-repair  &  & 0.93 & 0.95 & 0.90 & 0.94 & 0.02 \\
AI-repair  &  & 0.93 & 0.95 & 0.90 & 0.94 & 0.02 \\
\hline
Neurologist & \multirow{3}{*}{Focal slow} & 0.55 & 0.55 & 0.55 & 0.90 &  \\
AI-non-repair &  & 0.62 & 0.48 & 0.91 & 0.88 & 0.79 \\
AI-repair &  & 0.71 & 0.59 & 0.91 & 0.92 & 0.79 \\
\hline
\multicolumn{7}{p{230pt}}{GBS $=$ generalized background slowing, Focal slow $=$ focal slowing abnormalities.
P $=$ precision, R $=$ recall, ACC $=$ accuracy.}\\
\multicolumn{7}{p{230pt}}{
P-values are from McNemar's test comparing the accuracy of the AI system (with or without repair) to the neurologist's accuracy.}
\end{tabular}
\label{tab:comparison_ai_neurologist}
\end{table}

\subsection{Performance Comparison on TUAB Dataset}
Table \ref{tab:dataset_comparison} presents the confusion matrices and performance metrics for each dataset, along with the p-values obtained from Z-tests comparing the metrics between AI hybrid model's performance on the TUAB and our validation datasets.
The proposed model achieved F1 scores of 0.835 and 0.884, precisions of 0.827 and 0.857, recalls of 0.844 and 0.913, and accuracies of 0.884 and 0.890 on the TUAB and our validation datasets, respectively.
Z-tests conducted for each performance metric yielded p-values ranging from 0.59 to 0.76, all exceeding the 0.05 significance level. This indicates no statistically significant differences in model performance between datasets.
These results demonstrate the model's consistent performance and generalizability across different datasets, supporting its robustness and potential for real-world applications.
\begin{table}
\centering
\caption{Performance comparison of the proposed model on different datasets}
\begin{tabular}{lcccc}
\hline
\textbf{Dataset} & \textbf{TUAB} & \textbf{Our} & \textbf{P-value} \\
\hline
\textbf{Confusion Matrix} &
$\begin{bmatrix}
163 & 17\\
15 & 81
\end{bmatrix}$ &
$\begin{bmatrix}
47 & 7\\
4 & 42
\end{bmatrix}$ & - \\
\hline
F1 Score & 0.835 & 0.884 & 0.66 \\
Precision & 0.827 & 0.857 & 0.76 \\
Recall & 0.844 & 0.913 & 0.59 \\
Accuracy & 0.884 & 0.890 & 0.76 \\
\hline
\multicolumn{4}{p{230pt}}{TUAB $=$ The Temple University Abnormal EEG Corpus, Our $=$ validation dataset. P-values are from Z-tests comparing the performance metrics between the two datasets.}
\end{tabular}

\label{tab:dataset_comparison}
\end{table}

\begin{table}
    \caption{Accuracy Verification by 3 LLMs for AI-Generated EEG Reports}
    \centering
    \begin{tabular}{lccccc}
    \hline
    Category & Gwet AC1 (95\% CI) & F1 & ACC & P & R \\
    \hline
    GBS & 0.97 (0.95-0.98) & 1.00 & 1.00 & 1.00 & 1.00 \\
    Focal & 0.99 (0.98-1.0) & 1.00 & 1.00 & 1.00 & 1.00 \\
    \hline
    \multicolumn{6}{p{230pt}}{GBS $=$ generalized background slowing, Focal $=$ focal abnormalities, ACC $=$ accuracy, P $=$ precision, R $=$ recall}
    \end{tabular}    
    \label{tab:llm_verification}
\end{table}

\subsection{Accuracy of LLM Report Generation}

Table \ref{tab:llm_verification} shows the accuracy verification results of the AI-generated EEG reports by three LLMs. The inter-rater agreement among the three LLMs was assessed using Gwet's AC1 coefficient. For GBS, the AC1 coefficient was 0.97 (95\% CI: 0.95-0.98), and for focal abnormality, it was 0.99 (95\% CI: 0.98-1.0), indicating strong agreement in verifying report accuracy. Performance metrics for both GBS and focal abnormality showed F1 scores, precision, recall, and accuracy all at 1.0. These results confirm that the EEG reports generated by the hybrid AI algorithm and LLM were highly accurate and consistent across the three LLM models. Furthermore, these LLM-validated results were subsequently reviewed by human experts, who confirmed the 100\% accuracy of the generated reports, providing an additional layer of validation to the AI system's performance.

\section{Discussion}
\label{sec:discussion}
This study proposes an innovative hybrid artificial intelligence system for automatic interpretation of EEG background activity, addressing the clinical challenges faced by small hospitals and clinics that lack advanced EEG signal analysis systems and are prone to misinterpretation in manual EEG reading. In terms of PDR prediction, the system achieved a very high accuracy using only 1,530 EEG files as deep learning training data for PDR prediction. Additionally, it combines unsupervised artifact anomaly detection with expert guidelines and statistical analysis-based abnormality interpretation algorithms to precisely analyze EEG background features. The system utilizes open source programming environments, making it transferable across various operating systems and hardware platforms, which are all important innovative concepts in EEG interpretation.

One crucial algorithmic component of this system is the deep learning model used for PDR prediction. The best model, an ensemble approach, obtained an MAE of 0.237, an RMSE of 0.359, an accuracy of 91.8\% within a 0.6 Hz error, and an accuracy of 99\% within a 1.2 Hz error. Compared to traditional background PDR analysis methods \cite{PDR_Automated_EEG_Analysis_2011}, which reported an accuracy of 75.9\% within a 0.6 Hz error, 92.5\% within a 1.2 Hz error, and an RMSE of 0.96, our deep learning model demonstrated exceptional accuracy, particularly in the more stringent 0.6 Hz range. The superior performance of our approach can be attributed to several factors. Unlike traditional methods\cite{PDR_Automated_EEG_Analysis_2011} that rely on curve-fitting techniques or spectral analysis alone, our model leverages convolutional neural networks to learn complex patterns directly from the EEG data, capturing subtle features that may be missed by conventional methods.

For the evaluation of dataset size and reliability, our k-fold cross-validation results demonstrated no statistically significant differences between folds, indicating robust and consistent model performance across different data subsets. Furthermore, our smaller dataset validation experiments revealed that while performance generally improves with increased data, a dataset ratio of 0.4 to 0.6 of the original size can achieve reasonably accurate predictions. This corresponds to approximately 612 to 918 samples from our full dataset of 1,530. Based on these findings, we recommend a minimum training set size of 900 samples to achieve superior model accuracy. 

Regarding artifact removal, the unsupervised noise removal method based on the HBOS algorithm and neighboring electrode comparison effectively addressed the issue of EEG contamination by artifacts. There was no significant difference in PDR prediction performance between the artifact-repair and non-repair data. Moreover, in terms of feature generation accuracy, the artifact-repair group showed better accuracy and F1 scores compared to the non-repair group, although these differences were not statistically significant. This suggests that the unsupervised neighbor HBOS approach is a feasible method for artifact removal, reducing the need for manual labeling of artifacts.

The Hybrid AI model demonstrated competitive performance in detecting EEG abnormalities compared to human neurologists. For generalized background slowing (GBS), the model significantly outperformed neurologists (p $=$ 0.02) with higher accuracy and precision, but slightly lower recall. In focal slow abnormality detection, the AI using repair-group showed improvements across all metrics compared to neurologists, though these differences were not statistically significant (p $=$ 0.79). These findings suggest that AI-assisted EEG interpretation may complement human expertise, particularly in detecting generalized abnormalities, while highlighting areas for further improvement in focal abnormality detection.

Our Hybrid AI model also exhibited competitive performance compared to SCORE-AI \cite{JAMA_Neurology_2023}, while utilizing a significantly smaller training dataset. The model's accuracy for GBS (0.94) substantially exceeded SCORE-AI's accuracy for nonepileptiform-diffuse abnormalities (84.69\%). Similarly, for focal slow abnormalities, our model's accuracy (0.92) surpassed SCORE-AI's performance for nonepileptiform-focal abnormalities (85.71\%). Notably, these results were achieved with a training set of 1,530 labeled EEGs, compared to SCORE-AI's 30,493. This efficiency in data utilization, combined with our model's modular structure separating PDR prediction from abnormality detection, offers enhanced flexibility and adaptability. The superior performance, particularly in GBS detection, may be attributed to the synergy of deep learning for PDR prediction and expert-designed algorithms for abnormality detection, potentially offering greater versatility than SCORE-AI's single deep learning model approach.

Furthermore, our model's consistent performance across different datasets, including the public TUAB dataset, demonstrates its generalizability and potential for widespread clinical application. This is crucial for real-world implementation where EEG data may vary significantly between institutions. 

The use of large language models (LLMs) for generating report content demonstrated perfect accuracy in our experiment. LLMs can produce output in multiple languages, making them valuable for globalized healthcare settings where spelling or grammatical errors in medical records are common. However, LLMs may generate incorrect or non-existent content, known as AI hallucination\cite{largeAIHealthInformatics}, \cite{RN19}. In this study, 512 AI-generated text reports were repeatedly verified by AI, resulting in 100\% correct feature content with no inaccuracies, indicating the absence of hallucination. We observed that AI hallucination may arise from overly open-ended prompts, as seen in online chatbots. Additionally, providing insufficient input data or poorly structured input features to the LLMs can result in the generation of incorrect or inconsistent EEG reports. Improving prompting techniques is a viable solution for implementing AI in low-resource settings, as it reduces the need for expensive GPU hardware and large training datasets typically required for training complex AI models from scratch\cite{RN4}. We addressed this LLM shortcoming by enhancing our prompt engineering approach.

Limitations of this study include the primary training data being obtained from a single center, and the system's current focus on background interpretation, excluding seizure detection, photic stimulation, or hyperventilation interpretation. While we validated our model on the public TUAB dataset to assess generalizability, we were unable to conduct a head-to-head comparison to the existing state-of-the-art systems such as SCORE-AI due to no public access to the system and its validation dataset. Future research should incorporate data from multiple centers to further enhance the system's robustness across diverse clinical settings. Expanding the system's functionality could include adding EEG models for more EEG abnormalities. These enhancements would improve the system's clinical utility.

\section{Conclusion}
\label{sec:conclusion}
This study proposes a promising AI hybrid model for automatic interpretation of EEG background activity and report generation, providing an easily scalable and accurate solution for small hospitals and clinics. By combining AI techniques with expert-designed algorithms and directly generating text reports using LLMs, this system can efficiently assist neurologists in improving the diagnostic accuracy of EEG, reducing misdiagnosis rates, and simplifying the EEG interpretation process. Ultimately, it can enhance the quality of healthcare in resource-limited environments.

\section*{Acknowledgment} The authors would like to express their sincere gratitude to Dr. Che-Lung Su and Dr. Guan-Hsun Wang, neurologists at Min-Jhong Hospital, for their valuable contribution in reviewing and labeling the EEG data used in this study. The authors also extend their thanks to Professor Joseph Picone of Temple University for providing the TUAB abnormal EEG dataset, which was instrumental in validating the generalizability of our system.


\section*{References}

\begin{thebibliography}{99}

    \bibitem{RN2}
    J.W. Britton, L.C. Frey, and J.L. Hopp et al., ``Electroencephalography: An Introductory Text and Atlas of Normal and Abnormal Findings in Adults, Children, and Infants,'' American Epilepsy Society, pp. 48-50, 2016.
    
    \bibitem{RN11}
    M. Goldstein and A. Dengel, ``Histogram-based Outlier Score (HBOS): A fast Unsupervised Anomaly Detection Algorithm,'' in KI 2012: Advances in Artificial Intelligence, 2012, pp. 59-63.
    
    \bibitem{RN6}
    A. Gramfort, M. Luessi, E. Larson, D.A. Engemann, D. Strohmeier, C. Brodbeck, R. Goj, M. Jas, T. Brooks, L. Parkkonen, and M. Hämäläinen, ``MEG and EEG data analysis with MNE-Python,'' Frontiers in Neuroscience, vol. 7, 2013, doi: 10.3389/fnins.2013.00267.
    
    \bibitem{RN19}
    R. Hatem, B. Simmons, and J.E. Thornton, ``A Call to Address AI 'Hallucinations' and How Healthcare Professionals Can Mitigate Their Risks,'' Cureus, vol. 15, 2023.
    
    \bibitem{RN15}
    K. He, X. Zhang, S. Ren, and J. Sun, ``Deep residual learning for image recognition,'' in Proceedings of the IEEE conference on computer vision and pattern recognition, 2016, pp. 770-778.
    
    \bibitem{RN17}
    L.J. Hirsch et al., ``American Clinical Neurophysiology Society's Standardized Critical Care EEG Terminology: 2021 Version,'' Journal of Clinical Neurophysiology, vol. 38, no. 1, pp. 1-29, 2021, doi: 10.1097/wnp.0000000000000806.
    
    \bibitem{RN16}
    S.S. Lodder and M.J.A.M. van Putten, ``Quantification of the adult EEG background pattern,'' Clinical Neurophysiology, vol. 124, no. 2, pp. 228-237, 2013, doi: 10.1016/j.clinph.2012.07.007.
    
    \bibitem{RN1}
    M.R. Nuwer, ``Assessing digital and quantitative EEG in clinical settings,'' Journal of clinical neurophysiology, vol. 15, no. 6, pp. 458-463, 1998.
    
    \bibitem{RN5}
    R. Oostenveld, P. Fries, E. Maris, and J.M. Schoffelen, ``FieldTrip: Open Source Software for Advanced Analysis of MEG, EEG, and Invasive Electrophysiological Data,'' Computational Intelligence and Neuroscience, vol. 2011, pp. 1-9, 2011, doi: 10.1155/2011/156869.
    
    \bibitem{RN13}
    K. O'shea and R. Nash, ``An introduction to convolutional neural networks,'' arXiv preprint arXiv:1511.08458, 2015.
    
    \bibitem{RN10}
    S. Saba-Sadiya, E. Chantland, T. Alhanai, T. Liu, and M.M. Ghassemi, ``Unsupervised EEG Artifact Detection and Correction,'' Frontiers in Digital Health, vol. 2, 2021, doi: 10.3389/fdgth.2020.608920.
    
    \bibitem{PDR_Automated_EEG_Analysis_2011}
    S.S. Lodder and M.J.A.M. van Putten, ``Automated EEG analysis: Characterizing the posterior dominant rhythm,'' Journal of Neuroscience Methods, vol. 200, no. 1, pp. 86-93, 2011, doi: 10.1016/j.jneumeth.2011.06.008.
    
    \bibitem{RN14}
    C. Szegedy et al., ``Going deeper with convolutions,'' in Proceedings of the IEEE conference on computer vision and pattern recognition, 2015, pp. 1-9.
    
    \bibitem{RN9}
    W.O. Tatum IV, A.M. Husain, S.R. Benbadis, and P.W. Kaplan, ``Normal Adult EEG and Patterns of Uncertain Significance,'' Journal of Clinical Neurophysiology, vol. 23, no. 3, 2006, doi: 10.1097/01.wnp.0000220110.92126.a6.
    
    \bibitem{Multitaper_1982}
    D.J. Thomson, ``Spectrum estimation and harmonic analysis,'' Proceedings of the IEEE, vol. 70, no. 9, pp. 1055-1096, 1982, doi: 10.1109/proc.1982.12433.
    
    \bibitem{chollet2015keras}
    F. Chollet et al., ``Keras,'' 2015. [Online]. Available: https://keras.io
    
    \bibitem{JAMA_Neurology_2023}
    J. Tveit et al., ``Automated Interpretation of Clinical Electroencephalograms Using Artificial Intelligence,'' JAMA Neurology, vol. 80, no. 8, pp. 805, 2023, doi: 10.1001/jamaneurol.2023.1645.
    
    \bibitem{RN7}
    D. Yao, ``A method to standardize a reference of scalp EEG recordings to a point at infinity,'' Physiological measurement, vol. 22, no. 4, 2001, doi: 10.1088/0967-3334/22/4/305.
    
    \bibitem{RN8}
    D. Yao et al., ``Which Reference Should We Use for EEG and ERP practice?'' Brain Topography, vol. 32, no. 4, pp. 530-549, 2019, doi: 10.1007/s10548-019-00707-x.
    
    \bibitem{RN4}
    J. Zaghir, M. Naguib, M. Bjelogrlic, A. Névéol, X. Tannier, and C. Lovis, ``Prompt engineering paradigms for medical applications: scoping review and recommendations for better practices,'' arXiv preprint arXiv:2405.01249, 2024.
    
    \bibitem{geminiteam2024gemini}
    M. Reid et al., ``Gemini 1.5: Unlocking multimodal understanding across millions of tokens of context,'' arXiv preprint arXiv:2403.05530, 2024.
    
    \bibitem{anthropic_2024_introducing}
    Anthropic, ``Introducing the next generation of Claude,'' March 2024. [Online]. Available: https://www.anthropic.com/news/claude-3-family
    
    \bibitem{gpt4o_2024_introducing}
    OpenAI, ``Hello GPT-4o,'' May 2024. [Online]. Available: https://openai.com/index/hello-gpt-4o/
    
    \bibitem{MATLAB}
    MATLAB version R2023b, The MathWorks Inc., Natick, Massachusetts, United States, 2023.
    
    \bibitem{nuero_burden}
    V.L. Feigin et al., ``The global burden of neurological disorders: translating evidence into policy,'' Lancet Neurol, vol. 19, no. 3, pp. 255-265, 2020, doi: 10.1016/s1474-4422(19)30411-9.
    
    \bibitem{eeg_Significance}
    S. Siuly, Y. Li, and Y. Zhang, ``Significance of EEG Signals in Medical and Health Research,'' in Health Information Science: Springer International Publishing, 2016, pp. 23-41.
    
    \bibitem{remove_artifacts}
    X. Jiang, G.B. Bian, and Z. Tian, ``Removal of Artifacts from EEG Signals: A Review,'' Sensors, vol. 19, no. 5, pp. 987, 2019, doi: 10.3390/s19050987.
    
    \bibitem{Multitaper_2014}
    B. Babadi and E.N. Brown, ``A Review of Multitaper Spectral Analysis,'' IEEE Transactions on Biomedical Engineering, vol. 61, no. 5, pp. 1555-1564, 2014, doi: 10.1109/tbme.2014.2311996.
    
    \bibitem{Gwet_2008}
    K.L. Gwet, ``Computing inter-rater reliability and its variance in the presence of high agreement,'' British Journal of Mathematical and Statistical Psychology, vol. 61, no. 1, pp. 29-48, 2008, doi: 10.1348/000711006X126600.
    
    \bibitem{EEG_Abnormal}
    P.D. Emmady and A.C. Anilkumar, ``EEG Abnormal Waveforms,'' in StatPearls. Treasure Island (FL): StatPearls Publishing, 2023.
    
    \bibitem{eegDeepLearning}
    J. Zheng et al., ``Time-Frequency Analysis of Scalp EEG With Hilbert-Huang Transform and Deep Learning,'' IEEE Journal of Biomedical and Health Informatics, vol. 26, no. 4, pp. 1549-1559, 2022, doi: 10.1109/JBHI.2021.3110267.
    
    \bibitem{deepLearningStroke}
    P.J. Lin et al., ``A Transferable Deep Learning Prognosis Model for Predicting Stroke Patients' Recovery in Different Rehabilitation Trainings,'' IEEE Journal of Biomedical and Health Informatics, vol. 26, no. 12, pp. 6003-6011, 2022, doi: 10.1109/JBHI.2022.3205436.
    
    \bibitem{deepLearningBCI}
    X. Wu et al., ``Deep Learning With Convolutional Neural Networks for Motor Brain-Computer Interfaces Based on Stereo-Electroencephalography (SEEG),'' IEEE Journal of Biomedical and Health Informatics, vol. 27, no. 5, pp. 2387-2398, 2023, doi: 10.1109/JBHI.2023.3242262.
    
    \bibitem{deepLearningEpilepsy}
    Z. Shi, Z. Liao, and H. Tabata, ``Enhancing Performance of Convolutional Neural Network-Based Epileptic Electroencephalogram Diagnosis by Asymmetric Stochastic Resonance,'' IEEE Journal of Biomedical and Health Informatics, vol. 27, no. 9, pp. 4228-4239, 2023, doi: 10.1109/JBHI.2023.3282251.
    
    \bibitem{largeAIHealthInformatics}
    J. Qiu et al., ``Large AI Models in Health Informatics: Applications, Challenges, and the Future,'' IEEE Journal of Biomedical and Health Informatics, vol. 27, no. 12, pp. 6074-6087, 2023, doi: 10.1109/JBHI.2023.3316750.
    
    \bibitem{Lopez2017TUH}
    S. Lopez, ``Automated Identification of Abnormal EEGs,'' Master's thesis, Temple University, Philadelphia, PA, 2017.
    
    \end{thebibliography}
\end{document}